\documentclass[sigconf,10pt]{acmart}  

\usepackage{multirow}                   
\usepackage{enumitem}                   
\usepackage{balance}                    
\usepackage{adjustbox}
\usepackage{booktabs}
\usepackage{colortbl}
\usepackage{libertine}

\usepackage{amsmath}
\usepackage{algorithm}
\usepackage[noend]{algpseudocode}

\usepackage{graphicx}
\usepackage{float}
\usepackage{placeins}
\usepackage{subcaption}

\usepackage{pdfrender}

\newcommand{\myname}{Panopticus}

\newcommand{\system}{\myname}
\newcommand{\systembaseline}{\myname-Frame}

\graphicspath{{./figure/}}

\setcopyright{acmlicensed}
\copyrightyear{2024}
\acmYear{2024}
\acmConference[ACM MobiCom '24]{The 30th Annual International Conference on Mobile Computing and Networking}{November 18--22, 2024}{Washington D.C., DC, USA}
\acmBooktitle{The 30th Annual International Conference on Mobile Computing and Networking (ACM MobiCom '24), November 18--22, 2024, Washington D.C., DC, USA}
\acmDOI{10.1145/3636534.3690688}
\acmISBN{979-8-4007-0489-5/24/11}

\begin{document}

\title[Omnidirectional 3D Object Detection on Resource-constrained Edge Devices]{Panopticus: Omnidirectional 3D Object Detection on Resource-constrained Edge Devices}


\author{Jeho Lee}
\affiliation{%
  \institution{Yonsei University}
  \city{Seoul}
  \country{Republic of Korea}
}
\email{jeholee@yonsei.ac.kr}

\author{Chanyoung Jung}
\affiliation{%
  \institution{Yonsei University}
  \city{Seoul}
  \country{Republic of Korea}
}
\email{cy.jung@yonsei.ac.kr}

\author{Jiwon Kim}
\authornote{Current affiliation: Uppsala University, Sweden.}
\affiliation{%
  \institution{Yonsei University}
  \city{Seoul}
  \country{Republic of Korea}
}
\email{kim.j@yonsei.ac.kr}

\author{Hojung Cha}
\authornote{Corresponding author.}
\affiliation{%
  \institution{Yonsei University}
  \city{Seoul}
  \country{Republic of Korea}
}
\email{hjcha@yonsei.ac.kr}

\renewcommand{\shortauthors}{J. Lee, C. Jung, J. Kim, H. Cha}

\begin{abstract}
3D object detection with omnidirectional views enables safety-critical applications such as mobile robot navigation. Such applications increasingly operate on resource-constrained edge devices, facilitating reliable processing without privacy concerns or network delays. To enable cost-effective deployment, cameras have been widely adopted as a low-cost alternative to LiDAR sensors. However, the compute-intensive workload to achieve high performance of camera-based solutions remains challenging due to the computational limitations of edge devices. In this paper, we present \system{}, a carefully designed system for omnidirectional and camera-based 3D detection on edge devices. \system{} employs an adaptive multi-branch detection scheme that accounts for spatial complexities. To optimize the accuracy within latency limits, \system{} dynamically adjusts the model’s architecture and operations based on available edge resources and spatial characteristics. We implemented \system{} on three edge devices and conducted experiments across real-world environments based on the public self-driving dataset and our mobile 360° camera dataset. Experiment results showed that \system{} improves accuracy by 62\% on average given the strict latency objective of 33ms. Also, \system{} achieves a 2.1$\times$ latency reduction on average compared to baselines.
\vspace{-0.05in}
\end{abstract}


\vspace {-\baselineskip}
 


\maketitle


\section{Introduction}
\label{sec:intro}

\begin{figure}[t]
    \centering
    \includegraphics[width=\linewidth]{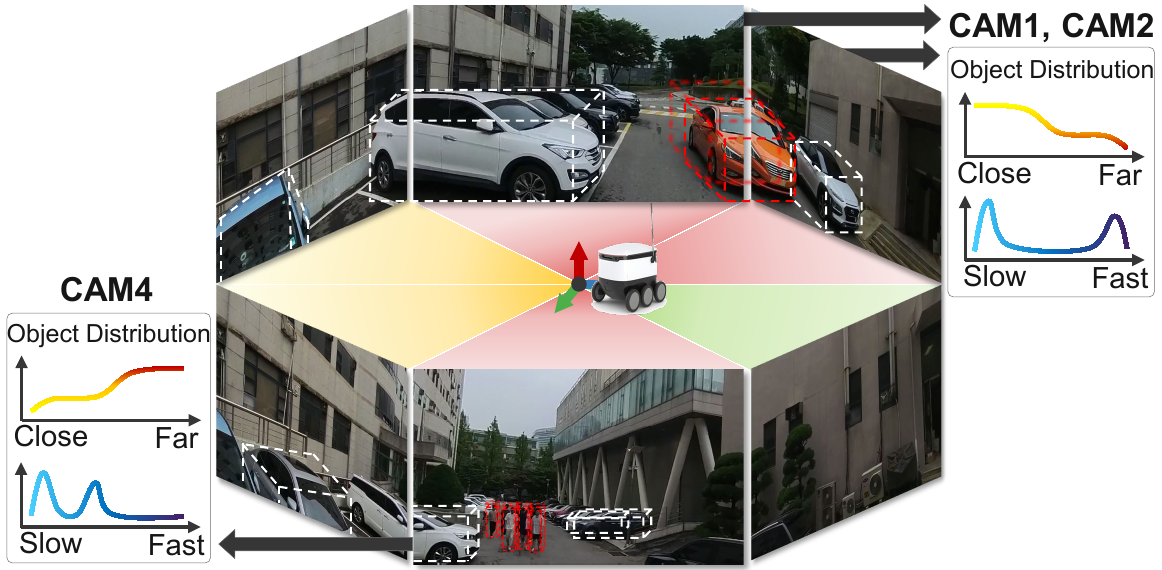}
    \caption{Obstacle avoidance based on omnidirectional 3D object detection. For optimized resource utilization, a robot prioritizes camera views with high spatial complexity (red) over those with low complexity (green and yellow).}
    \label{fig:figure1}
    \vspace {-0.2in}
\end{figure}

Along with the advances in computer vision and deep neural networks (DNNs), 3D object detection has become a core component of numerous applications. For example, autonomous vehicles rely on precise and real-time perception of objects in an environment to establish safe navigation routes~\cite{surveyautodrive}. Since objects can approach from any direction, as shown in Figure ~\ref{fig:figure1}, it is crucial to ensure perception through a comprehensive 360° field of view (FOV). Such omnidirectional perception requires the processing of substantial amounts of sensor data and demands high-end computing devices with AI accelerators for real-time processing~\cite{teslachip}.

Recently, the demand for mobile applications using omnidirectional 3D object detection has become widespread. Robots or drones providing personal services such as surveillance can benefit from such technology~\cite{Floreano2015}. In addition, detecting surrounding obstacles and providing audible warnings of potential hazards can help people with visual impairments~\cite{vis_impair_usecase1, vis_impair_usecase2}. These personalized applications must be processed on an edge device to minimize user privacy issues or network overheads. However, even the latest NVIDIA Jetson Orin series~\cite{nvidia_jetson_orin}, offering advanced edge compute power, has 6.7$\times$ to 13.5$\times$ fewer Tensor cores for AI acceleration compared to the powerful A100~\cite{nvidia_a100} used for cloud computing, which has the same underlying GPU architecture. Furthermore, edge AI applications must consider practical factors such as cost-effective deployments. As a result, much effort has been made to support such applications with low-cost cameras~\cite{lowcost_camera_usecase1, lowcost_camera_usecase2, kinematic3D_eccv, lift_splat_shoot}. Specifically, multiple cameras or a mobile 360° camera are utilized to facilitate omnidirectional perception~\cite{omni_camera_percept_example1, omni_camera_percept_example2, omni_camera_percept_example3}.

Edge AI services have a wide spectrum of accuracy and latency requirements. Despite recent advances, prior works have limitations in supporting both efficiency and accuracy on resource-constrained edge devices. DeepMix~\cite{DeepMix} offloaded complex DNN-based object detection tasks to a cloud server to reduce the computational burden on an edge device. Offloading omnidirectional perception tasks, however, may cause significant edge-cloud communication latency due to massive data transmission. PointSplit~\cite{PointSplit} supports parallelized operation on edge GPU and NPU, but the scheme is optimized for a specific 3D detection pipeline utilizing an RGB-D sensor with limited FOV. Meanwhile, various methods~\cite{kinematic3D_eccv, lift_splat_shoot, MonoCon, BEVFormer} have enhanced the accuracy of camera-based solutions, which pose inherent difficulties due to the absence of 3D depth information. A line of works~\cite{BEVDepth, BEVStereo, BEVHeight} has focused on developing DNNs to enhance depth prediction from RGB images. Also, the adoption of large-scale DNNs, such as feature extraction backbones using high-resolution images, is essential for accuracy improvement~\cite{BEVFormer_v2}. However, processing multiple compute-intensive DNN tasks with omnidirectional inputs places substantial computational demands on resource-constrained edge devices.

In this paper, we propose \system{}, a system that maximizes the accuracy of omnidirectional 3D object detection while meeting the latency requirements on edge devices. We preliminarily observed that camera-based 3D detectors have varying detection capabilities depending on spatial characteristics, which are determined by various factors such as the number or movement of objects. The key idea of \system{} is to process each camera view optimally based on the understanding of short-term dynamics in spatial distribution. For example, a camera view containing a few static and proximate objects can be processed with a lightweight inference configuration to reduce the latency with a minimal accuracy loss. The saved latency margin can then be utilized to assign a high-performing inference configuration to a complex view where objects are moving fast or in a distant location, as shown in Figure ~\ref{fig:figure1}.

Several challenges exist in the design of \system{}. First, prior 3D detection models fail to provide an efficient and dynamic inference scheme capable of differentiating the inference configuration for each camera view in the same video frame, such as backbone capacity or the use of enhanced depth estimation. Additionally, the model’s architecture must be adjustable to accommodate the various constraints, such as latency requirements, on a given device. Second, to maximize the accuracy within latency requirements, the optimal inference configuration must be decided for each camera view. This requires a runtime analysis of both changes in spatial distribution and the expected performance of inference configurations.

To enable architectural and operational adjustments of the model, we introduce an omnidirectional 3D object detection model with multiple inference branches. The model processes each view using one of the branches with varying detection capabilities, enabling fine-grained utilization of edge computing resources. The model’s architecture is designed to be modular, enabling flexible deployments by detaching a branch that violates given constraints. For the second challenge of maximizing accuracy within latency limits, we introduce a spatial-adaptive execution scheme. At runtime, the scheme predicts the performance of each branch based on the expected spatial distribution of the surrounding objects. Optimal combinations of branches and camera views, which maximize overall estimated accuracy while meeting the latency goal, are then selected for inference.

We implemented \system{} on three edge devices with different computational capabilities. The system was evaluated in various real-world environments, such as urban roads and streets, using a public autonomous driving dataset and our custom mobile 360° camera testbed. Extensive experiments showed that \system{} outperformed its baselines under diverse scenarios in terms of both detection accuracy and efficiency. 

The key contributions of our work are as follows:

\begin{itemize}[]

    \item To the best of our knowledge, \system{} is the first omnidirectional and camera-based 3D object detection system that achieves both accuracy and latency optimization on resource-constrained edge devices.

    \item We conducted an in-depth study to explore the varying capabilities of recent 3D detectors influenced by diverse characteristics of objects and spaces. \system{} provides fine-grained control over omnidirectional perception and edge resource utilization, adapting to varying spatial complexities in dynamic environments.

    \item We fully implemented \system{} as an end-to-end edge computing system using both a public self-driving dataset and our mobile 360° camera testbed, showcasing its adaptability to the resource constraints of edge devices across a range of real-world conditions.
    
\end{itemize}

\section{Background: Omnidirectional 3D Object Detection}
\label{sec:back}

3D object detection aims to identify objects in space and predict their properties such as 3D location, size, and velocity. The predicted object information is utilized by application functionalities such as obstacle avoidance for robot navigation. Safe navigation cannot be solely ensured by Simultaneous Localization and Mapping (SLAM), lacking the ability to model the object sizes or movements in real-time. A robot must plan its navigation path based on obstacles' location and size, or even their predictive trajectory, to prevent collisions beforehand. Moreover, in complex outdoor environments where objects can approach from multiple directions, the ability to detect surrounding objects becomes essential.

\begin{figure}[t]
    \centering
    \includegraphics[width=\linewidth]{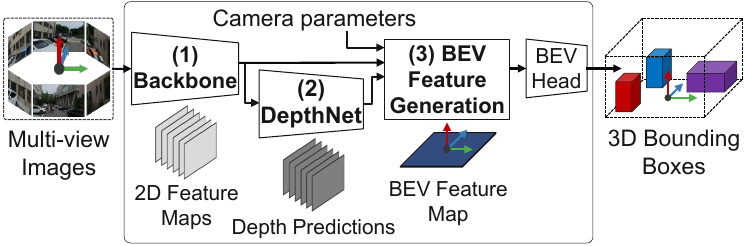}
    \caption{Baseline pipeline of BEV-based 3D detection with camera images.}
    \label{fig:figure2}
    \vspace {-0.2in}
\end{figure}

Existing methods for omnidirectional 3D object detection utilize LiDAR sensors or multiple cameras providing a 360° perception range. While LiDAR sensors offer accurate object localization based on depth measurements, the camera-based solutions have recently drawn attention due to their cost-effectiveness. Recent camera-based detectors aggregate information from multiple camera images into a bird’s-eye-view (BEV) space, providing a top-down representation of the surrounding 3D space. Early work~\cite{lift_splat_shoot} proposed an end-to-end trainable method to extract BEV features directly from multi-view images. Building upon~\cite{lift_splat_shoot}, BEVDet~\cite{huang2021bevdet} enables the detection of surrounding 3D objects using the extracted BEV features. Due to its simplified and scalable architecture, many of the latest BEV-based 3D detectors~\cite{BEVDepth, huang2022bevdet4d, BEVStereo, BEVHeight} followed BEVDet’s inference pipeline, as shown in Figure ~\ref{fig:figure2}. Such detectors have overcome the monocular ambiguity of camera-based approaches by introducing enhanced methods for each stage of the baseline BEVDet pipeline, achieving accuracy comparable to the LiDAR counterpart~\cite{centerpoint}. Table ~\ref{tab:table1} lists these methods, which are described in the following.

The first stage involves extracting 2D feature maps from multi-view images using backbone neural networks, such as ResNet~\cite{resnet}, widely used in vision tasks. It is well-known that increasing the backbone capacity, e.g., the number of layers, or input image resolution leads to accuracy improvements. For example, combining a 152-layer ResNet with a 720$\times$1,280 (height$\times$width) resolution yields a more detailed feature map than a 34-layer ResNet with 256$\times$448 resolution, thereby enhancing the detection of small and distant objects. The second stage of BEV-based detection is to transform the extracted image features into 3D space using predicted depth. For this stage, prior work~\cite{BEVDepth} employed a depth estimation neural network, i.e., DepthNet, supervised by dense depth data generated from 3D point clouds. Accurate metric depth predictions (in meters) from images allow the detector to better distinguish objects from the backgrounds. The third stage involves projecting features scattered in each 3D camera coordinate into a unified BEV grid using camera parameters, generating a BEV feature map. Recent works~\cite{huang2022bevdet4d, BEVFormer} have proposed techniques that fuse the BEV feature map from a previous frame with the feature map of the current frame. By exploiting temporal cues, the technique improves perception robustness, enabling the detection of temporarily occluded objects and the accurate prediction of object velocities. Lastly, the neural networks in the BEV head generate 3D bounding boxes and their properties, e.g., location and velocity, using the BEV features.

\begin{table}[t]
\caption{Design space of BEV-based 3D detection.}
\label{tab:table1}
\centering
\begin{adjustbox}{width=0.82\columnwidth,center}
\begin{tabular}{cc}
\hline\hline
\textbf{Stage} & \textbf{Method} \\ 
\hline
\multirow{4}{*}{\begin{tabular}[c]{@{}c@{}}(1) 2D Feature Extraction \\ Backbone Neural Network \\ (Input Image Resolution)\end{tabular}}
& ResNet34 (256$\times$448) \\
& ResNet50 (400$\times$704) \\
& ResNet101 (544$\times$960) \\
& ResNet152 (720$\times$1,280) \\ 
\hline
\multirow{2}{*}{\begin{tabular}[c]{@{}c@{}}(2) Depth Prediction \\ Neural Network\end{tabular}}
& Sparse DepthNet~\cite{huang2021bevdet} \\
& Dense DepthNet~\cite{BEVDepth} \\ 
\hline
(3) BEV Feature Generation 
& Temporal BEV Fusion~\cite{huang2022bevdet4d} \\ 
\hline\hline
\vspace {-0.2in}
\end{tabular}
\end{adjustbox}
\end{table}

\section{Preliminary Experiment}
\label{sec:preliminary}

We present in-depth observations on methods to improve the accuracy of the baseline BEVDet model mentioned in Section ~\ref{sec:back}. Based on the results, we derived key insights and challenges in the design of \system{}.

\subsection{Experiment Setup}
\label{sec:preliminary-setup}

Based on the baseline BEVDet model, we established 16 variants with different detection capabilities by combining the design parameters in Table ~\ref{tab:table1}. Each variant utilizes one of four combinations of ResNet backbones and input resolutions listed in Table ~\ref{tab:table1}, such as ResNet34 backbone with a resolution of 256$\times$448, which we refer to as BEVDet-R34. Also, each variant incorporates one of two DepthNets, differing in depth prediction capabilities. Last, a group of variants utilizes a temporal fusion technique, further enhancing BEV feature generation. To compare with the state-of-the-art LiDAR-based method, we profiled the CenterPoint~\cite{centerpoint} model. For model training, we utilized the nuScenes dataset~\cite{nuscenes} with ground-truth annotated 850 scenes, each of which contains a 20-second duration of video frames. Every frame provides 360° horizontal FOV through six camera images, each with a resolution of 900$\times$1,600. For evaluation metrics, we used average precision (AP), widely employed to assess detection accuracy, and true positive (TP) errors. TP errors include average translation error (ATE) in meters and average velocity error (AVE) in meters per second, which represent the prediction errors of 3D location and speed for correctly detected objects, respectively. All metrics are calculated for each object type, such as cars and bicycles, and averaged across all types to report the model's final detection performance, i.e., mAP, mATE, and mAVE. Inference latency was profiled on a recent edge device, Jetson AGX Orin~\cite{nvidia_jetson_orin} with a CUDA~\cite{nvidia_cuda_toolkit} GPU.

\begin{figure}[t]
    \centering
    \includegraphics[width=0.9\linewidth]{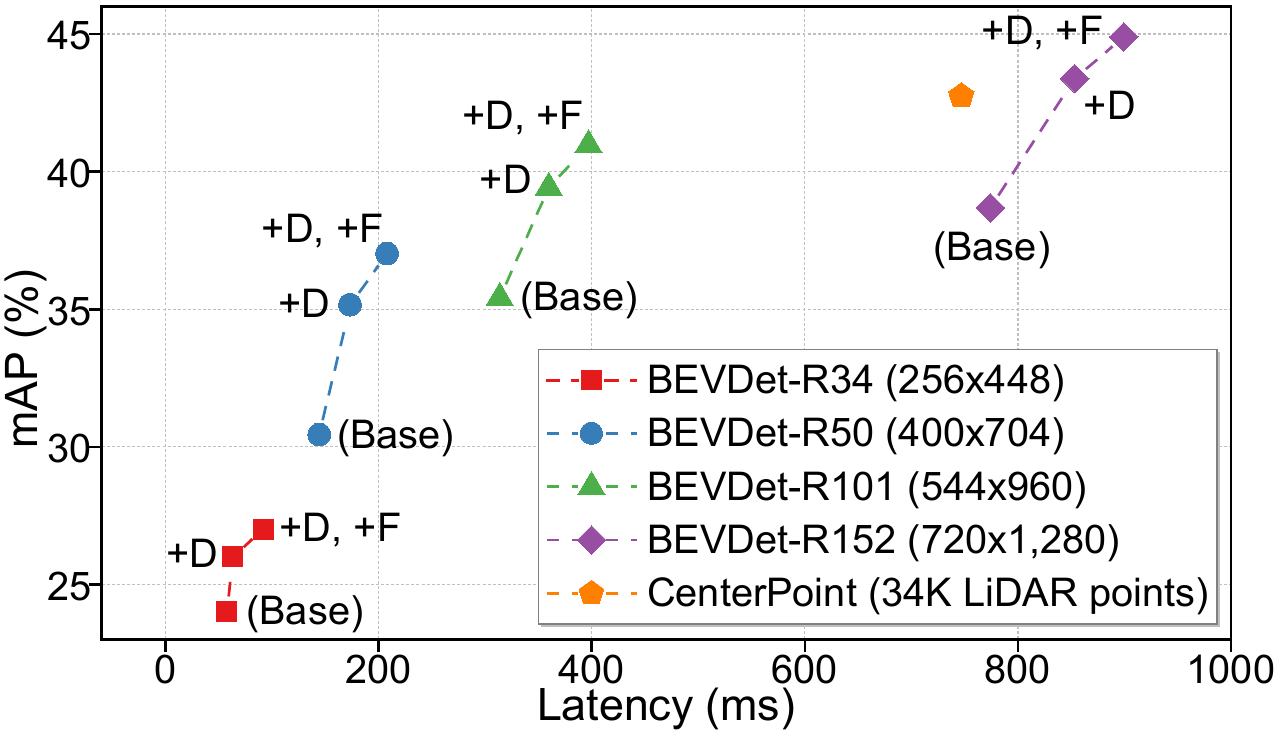}
    \caption{Detection accuracy and latency of BEVDet variants and LiDAR-based detector with edge GPU. +D and +F denote the incorporation of dense DepthNet and temporal fusion, respectively.}
    \label{fig:figure3}
    \vspace {-0.2in}
\end{figure}

\subsection{Observations}
\label{sec:preliminary-observations}

\textbf{Accuracy-latency trade-offs in different models.} Both detection accuracy and latency are crucial factors in building a 3D detection system. As shown in Figure ~\ref{fig:figure3}, the profiled 3D detection models exhibit various accuracy-latency trade-offs. Increasing both backbone capacity and input resolution clearly enhances mAP, but leads to high latency. For example, BEVDet-R152, which has the largest backbone capacity and resolution, achieves a 1.6$\times$ mAP improvement compared to BEVDet-R34. However, the inference latency is substantially increased by +717ms. Incorporating a dense DepthNet and temporal feature fusion into BEVDet-R152 further improves mAP by 12.1\% and 16.1\%, respectively, outperforming LiDAR-based CenterPoint. However, employing all of these methods to enhance camera-based detection can result in considerable inference latency on resource-constrained edge devices.

\begin{figure}[t]
    \centering
    \includegraphics[width=\linewidth]{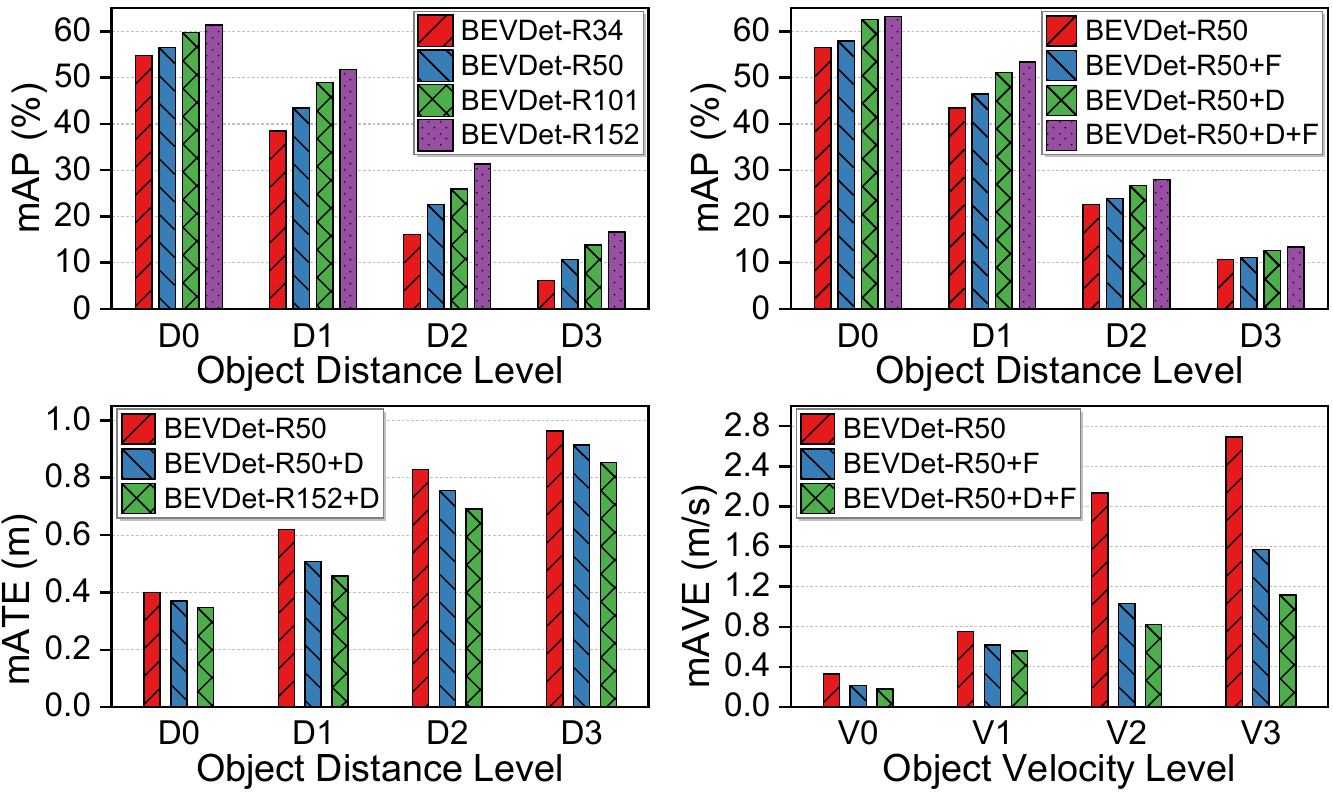}
    \caption{mAP (first row) and TP errors (second row) over objects with different distances and velocities. +D and +F denote the incorporation of dense DepthNet and temporal fusion, respectively.}
    \label{fig:figure4}
    \vspace {-0.2in}
\end{figure}

\noindent\textbf{Detection performance over various object properties.} Real-world outdoor scenes encounter objects with diverse characteristics such as moving speed. We categorized objects based on the property and its levels defined in Table ~\ref{tab:table2}. For brevity, we focused on the distance between objects and the camera, as well as object velocity. As shown in the first row of Figure ~\ref{fig:figure4}, all models achieve high mAP for objects with a D0 distance level. However, as the object distance level increases, lightweight models show significantly lower mAP. For example, the mAP of BEVDet-R34 for D3 objects is 2.7$\times$ lower than that of BEVDet-R152. The increase in errors of location and velocity predictions for distant and fast-moving objects, as shown in the second row of Figure ~\ref{fig:figure4}, is mitigated in heavier models. Specifically, leveraging both temporal fusion and dense DepthNet reduces velocity prediction error for V3 objects by 2.4$\times$ compared to BEVDet-R50. Reducing the prediction errors of object properties is important for safety-critical applications. For instance, autonomous robots plan a navigation path based on the understanding of surrounding objects’ positions and movements. In summary, the detection capabilities of models vary across diverse object properties, requiring the use of appropriate models with an awareness of objects’ characteristics.

\begin{table}[t]
\caption{3D object categorization.}
\label{tab:table2}
\centering
\begin{adjustbox}{width=\columnwidth,center}
\begin{tabular}{cc}
\hline\hline
\textbf{Property (unit)} & \textbf{Range (level)} \\
\hline
\multirow{2}{*}{\vspace*{3mm}Distance (\textit{$m$})\vspace*{3mm}}
& \begin{tabular}{ccccc}
     0--10 & 10--20 & 20--30 & 30--40 & 40 $\uparrow$  \\
     (D0) & (D1) & (D2) & (D3) & (D4) \end{tabular} \\
\hline
Velocity (\textit{$m/s$}) & \begin{tabular}{cccc}
     0--0.2 (V0) & 0.2--1 (V1) & 1--5 (V2) & 5 $\uparrow$ (V3) \end{tabular} \\
\hline
Size (\textit{$m^3$}) & \begin{tabular}{cccc}
     0--1 (S0) & 1--5 (S1) & 5--15 (S2) & 15 $\uparrow$ (S3) \end{tabular} \\
\hline\hline
\vspace {-0.2in}
\end{tabular}
\end{adjustbox}
\end{table}

\begin{figure}[t]
    \centering
    \begin{subfigure}[t]{\linewidth}
        \centering
        \includegraphics[width=\columnwidth]{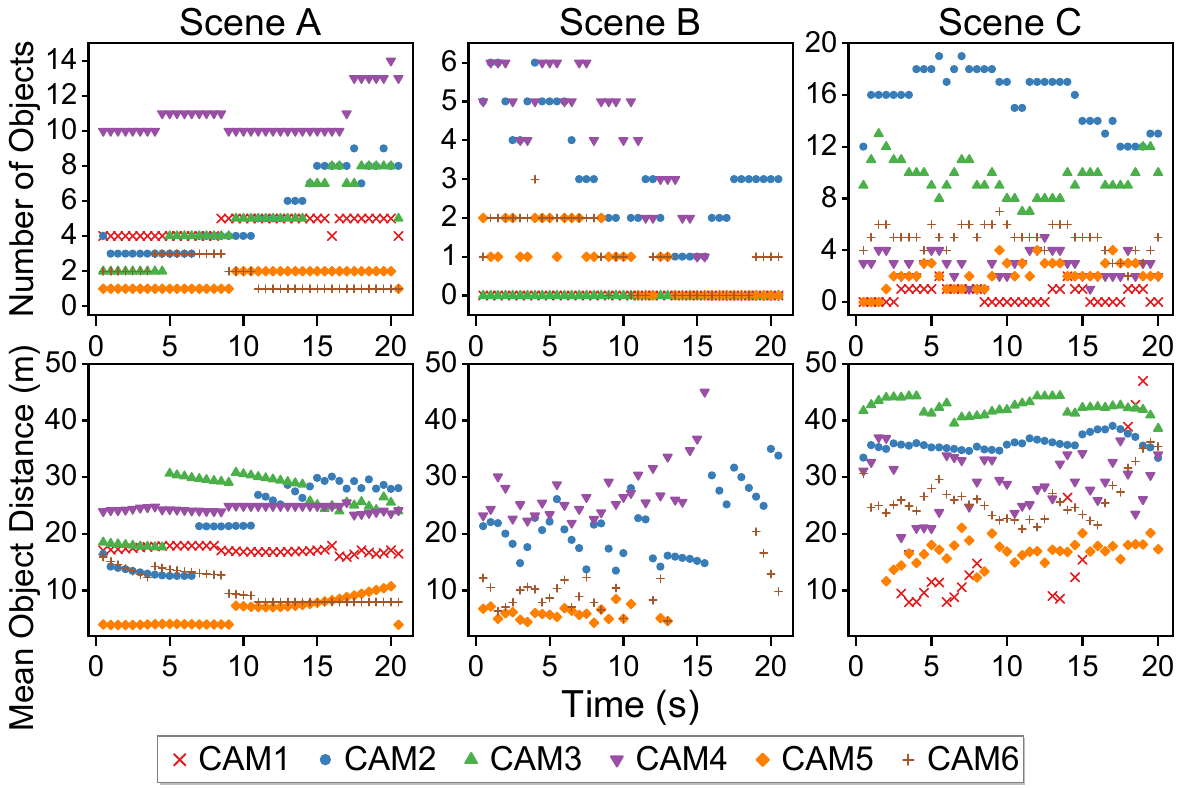}
        \subcaption{Object distribution across camera views}
        \label{fig:figure5(a)}
    \end{subfigure}
    \begin{subfigure}[t]{\linewidth}
        \centering
        \includegraphics[width=\columnwidth]{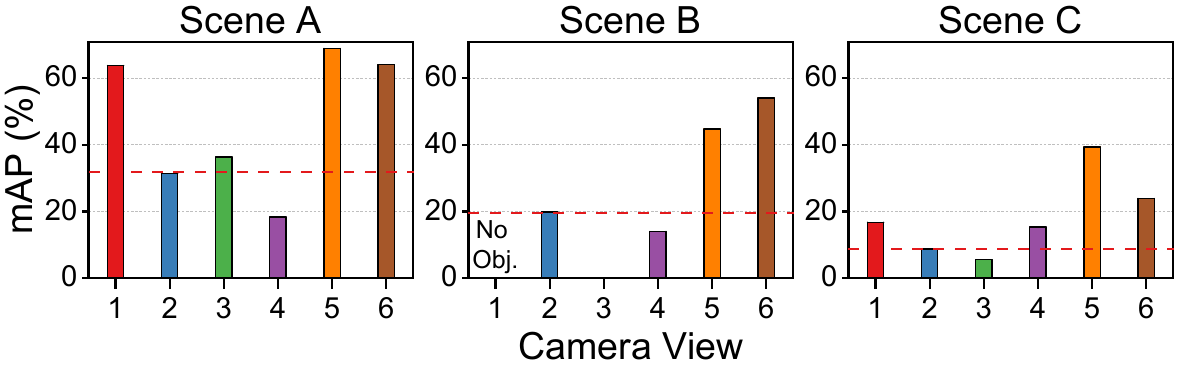}
        \subcaption{mAP of BEVDet-R50 for different camera views. Dashed line indicates overall mAP across camera views}
        \label{fig:figure5(b)}
    \end{subfigure}
    \caption{Distribution and mAP in example scenes.}
    \label{fig:figure5}
    \vspace {-0.2in}
\end{figure}

\noindent\textbf{Detection performance over diverse spatial distributions.} We examined how detection performance varies depending on spatial characteristics. Figure ~\ref{fig:figure5} shows the spatial object distribution and mAP in three example scenes from the nuScenes dataset. For each camera view (e.g., CAM1), we analyzed the spatial distribution by calculating both the number of objects and their average distance from the camera. Figure ~\ref{fig:figure5(a)} shows that the overall scene complexity increases from Scene A to C, considering distribution in all camera views. Therefore, as shown in Figure ~\ref{fig:figure5(b)}, the overall mAP gradually decreases. Note that the relative difference in distribution and mAP is even more significant across camera views. In Scene A, CAM4, which captures many distant objects, has the lowest mAP, whereas camera views with a smaller number of closer objects, such as CAM5, have a higher mAP. Meanwhile, in Scene C, the mAP between CAM3 and CAM5 differs by as much as 7.1$\times$, indicating significant differences in spatial complexities. Consequently, diversity in spatial distribution across scenes and camera views necessitates model selection for \textit{each view} to achieve the best performance.

\subsection{Summary and Challenges}
\label{sec:preliminary-summaryandchallenge}

Based on the observations, we derived insights to enhance both the accuracy and efficiency of omnidirectional 3D detection. Prior BEV-based models have made advancements in camera-based detection accuracy, but may result in substantial computational demands for a resource-constrained device. We observed that the models’ detection capabilities differ by object properties such as distance and velocity. Moreover, the variance in spatial distribution across scenes, even among different camera views within the same scene, suggests a need for employing diverse models to enhance overall performance. The experimental results inspired us to explore a high-accuracy omnidirectional perception scheme tailored for limited edge resources. This scheme would utilize the appropriate models for multi-view images, considering the models' capabilities and the spatial distribution variations across camera views.

However, realizing this scheme poses several challenges. First, previous BEV-based models lack dynamic inference capabilities to process each camera view with varying inference configurations, such as the choice of backbone or DepthNet types. An approach to load all possible models into memory and assign them across views~\cite{remix} may incur significant memory and latency overheads due to the redundant components among models. Moreover, the architecture of the BEV-based model needs to be adjustable considering various constraints such as device capabilities and latency requirements. Second, to maximize detection accuracy while meeting the latency requirements, the optimal inference configurations for all camera views must be established at each video frame. This requires real-time analysis of (1) changes in the surrounding spatial distribution within a dynamic environment and (2) the expected performance of different inference configurations.

\section{Overview of \system{}}
\label{sec:overview}

\begin{figure}[t]
    \centering
    \includegraphics[width=\linewidth]{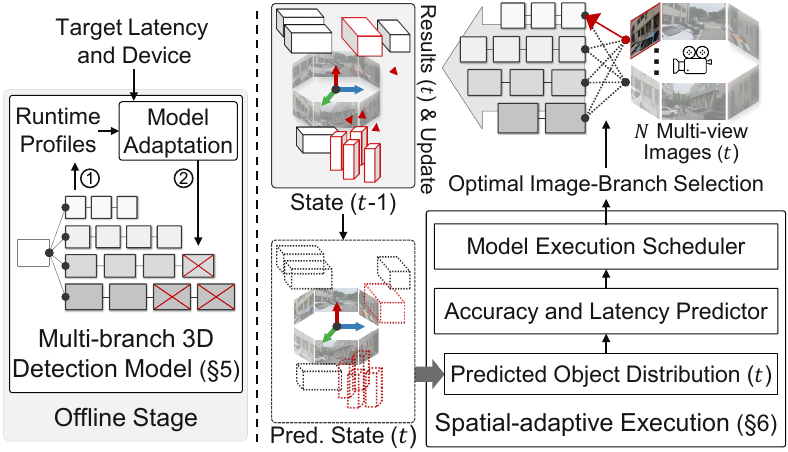}
    \caption{System architecture of \system{}.}
    \label{fig:figure6}
    \vspace {-0.2in}
\end{figure}

\begin{figure*}[t]
    \centering
    \includegraphics[width=0.95\textwidth]{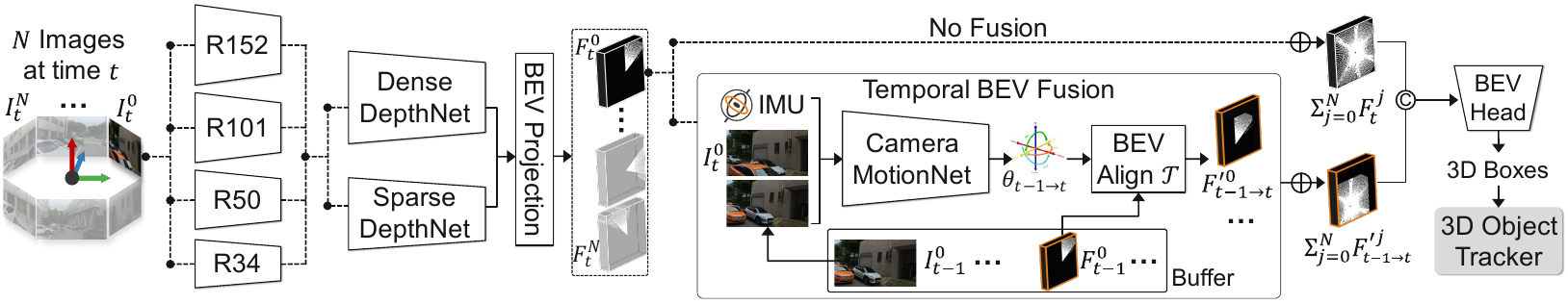}
    \caption{Architecture of multi-branch omnidirectional 3D object detection model. Dashed lines are the possible execution paths for each camera view.}
    \label{fig:figure7}
    \vspace {-0.2in}
\end{figure*}

Motivated by our observations and insights, we present \system{}, an omnidirectional 3D object detection system designed for resource-constrained edge devices. \system{} aims to maximize detection accuracy while meeting the latency objective with a given edge device. Figure ~\ref{fig:figure6} illustrates the system architecture of \system{}. 

To accommodate camera views with varying inference requirements, we propose an omnidirectional 3D detection model equipped with multiple inference branches. Each branch comprises different modules from BEV-based 3D detectors, providing flexibility in processing each camera view with distinct detection capabilities. The model architecture can be modified by simply detaching the branch modules to meet the diverse requirements accounting for latency and device. In the offline stage, runtime characteristics such as inference latency are profiled for each branch. Modules within a heavy branch that exceed latency or memory constraints are removed. At runtime, each of $N$ multi-view images at time $t$ is processed by the proper branch to generate accurate 3D bounding boxes. The optimal selection of branch and image combinations is determined by a spatial-adaptive execution scheduler. The scheduler estimates the expected accuracy and inference time for each branch-image pair, based on the predicted spatial distribution for the incoming frame. Then, the optimal $N$ image-branch pairs are selected, which maximizes overall accuracy while ensuring the total predicted latency is within the target latency. The detected boxes are used to update the tracked status of the surrounding objects, allowing the continuous observation of spatial characteristics.

\section{Multi-branch Omnidirectional 3D Object Detection}
\label{sec:system_main_1}

We describe a multi-branch design of our omnidirectional 3D detection model. We also explain how the model is modified to fit the given device and latency target.


\subsection{Model Design}
\label{sec:multibranch_model_design}

The goal of our multi-branch model is to allow each region in a surrounding space can be processed through various capabilities of the BEV-based 3D detectors detailed in Table ~\ref{tab:table1}. Instead of simply using all types of models loaded on the memory, we modularized the models and combined the core modules, eliminating redundant ones. The model follows the baseline pipeline of the BEV-based 3D detection illustrated in Figure ~\ref{fig:figure2}. However, as shown in Figure ~\ref{fig:figure7}, an image can be processed by one of several modules at each inference stage. Each branch within the model represents a unique execution path for an image, comprising a sequence of module decisions at all stages. We detailed the branch design following the order of the baseline pipeline.

\begin{figure}[t]
    \centering
    \includegraphics[width=0.8\linewidth]{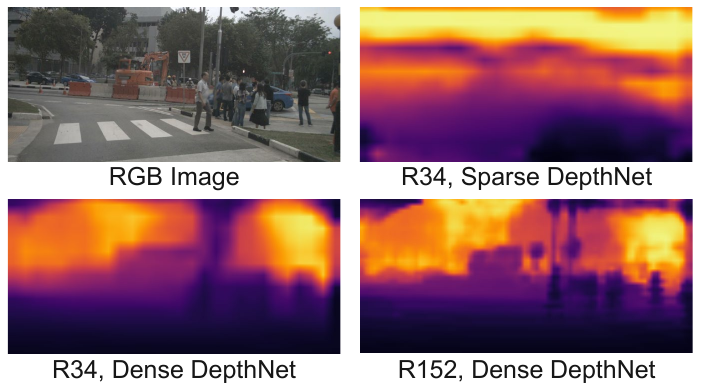}
    \caption{Predicted depth maps from RGB image.}
    \label{fig:figure8}
    \vspace {-0.2in}
\end{figure}

The initial stage of BEV feature generation is to extract 2D feature maps from all multi-view images. At this stage, each image can be processed through one of four backbone networks, each of which is differentiated by its capacity and input resolution, as detailed in Table ~\ref{tab:table1}. From the extracted 2D feature maps, the model infers the 3D depth information in each camera coordinate. To enable the adjustment of quality and latency of the depth prediction, our model is equipped with two types of DepthNets. One is a simple 1-layer convolutional network trained using only object depth loss, providing sparse supervision signals. The other is a deep convolutional network trained on dense depth maps derived from 3D point clouds. As illustrated in Figure ~\ref{fig:figure8}, the densely supervised DepthNet generates an accurate and detailed depth map from an image. The depth prediction quality is further increased by incorporating a large-scale backbone network with high-resolution images, providing a clear distinction between foreground objects and the background.

BEV feature maps are generated by projecting the extracted 2D feature map into a unified grid space using the predicted depth and camera parameters. Fusing the generated BEV features over time enhances the robustness of perception by providing contiguous observations of objects. As shown in Figure ~\ref{fig:figure7}, we developed a module that can fuse the BEV features extracted from consecutive images of each camera view $j$. Due to the possible movements of the camera (or a robot carrying the camera), directly fusing the BEV features may result in misalignment of spatial features over time, greatly reducing the accuracy gain. Therefore, the module aligns the last frame's BEV feature map $F_{t-1}^j$ into the current frame’s camera coordinate system. The aligned BEV feature map $F_{t-1 \rightarrow t}^{\prime j}$ is acquired as follows:
\begin{equation}
F_{t-1 \rightarrow t}^{\prime j} = \mathcal{T} ( F_{t-1}^{j}, \theta_{t-1 \rightarrow t}^{-1} ),
\label{equ:equation1}
\end{equation}
where $\mathcal{T}$ is the operation of transforming the spatial locations of features in the BEV grid using the relative camera pose, i.e., the camera motion $\theta_{t-1 \rightarrow t}$. To estimate $\theta_{t-1 \rightarrow t}$, we employ a neural network that uses inertial sensor data and consecutive images. At runtime, \system{} buffers the previous image $I_{t-1}^j$ and $F_{t-1}^j$ of each camera view $j$. Since the prediction results of $\theta_{t-1 \rightarrow t}$ are consistent across camera views, the model predicts it once per frame, allowing the sharing of predicted $\theta_{t-1 \rightarrow t}$ with the alignment operations for other views.

After aggregating $F_{t}^j$ and the aligned $F_{t-1 \rightarrow t}^{\prime j}$ across all camera views, the model concatenates the integrated BEV feature maps, which are then fed into the BEV head. The resulting 3D bounding boxes are post-processed using non-max suppression (NMS) to remove duplicates.

\system{} utilizes a 3D object tracker to keep track of the latest states of detected objects. We employed a 3D Kalman filter that can efficiently forecast and update the status of tracked objects. Our model features a branch that outputs the predicted states of tracked objects, skipping new detection for the target camera view. The future state $\mathrm{x}_t^\prime$ of a tracked object at the incoming time $t$ is predicted as follows:
\begin{equation}
\mathrm{x}_t^\prime = A_{t-1} \cdot \mathrm{x}_{t-1},
\label{equ:equation2}
\end{equation}
where $A_{t-1}$ and $\mathrm{x}_{t-1}$ are the state transition model and the object state vector at time $t-1$, respectively. $\mathrm{x}_{t-1}$ is parameterized by the object’s 3D location $(x,y,z)$, velocity $(v_x,v_y,v_z)$ and size $(w,h,l)$. Future state estimation of the target object using $A_{t-1}$ is simply applying the velocity predicted by the BEV head to its location, which is processed instantaneously. Overall, the image from each camera view can be processed by one of 17 branches: 16 detection branches and 1 lightweight tracker's branch utilizing objects' predicted states.

\subsection{Model Adaptation}
\label{sec:multibranch_model_adaptation}

\system{} supports offline model adaptation to meet the memory and latency constraints on a given device. To achieve this, the memory consumption and processing latency of all modules in the model are profiled on the target device. First, the model is adjusted to adhere to the memory constraints. For example, for the Jetson Orin Nano~\cite{nvidia_jetson_orin} with 6.3 GB of limited memory, large-sized modules such as the R152 backbone are detached to prevent overloading the memory capacity. Second, some inference branches require significant computation; in fact, inferring a single image may surpass the target latency. To meet latency requirements, branches with latency profiles that exceed the target limit are also removed. The model, with its number of branches $M$ changed, is then deployed on the target device.

\section{Spatial-adaptive Execution}
\label{sec:system_main_2}

\system{} dynamically schedules the inference configuration of the multi-branch model at runtime to maximize detection accuracy within the latency target $T$. Specifically, the goal of the scheduler is to find the optimal selection $c^*$ for $N$ image-branch pairs, given the possible inference branches $\{B^i\}_{i=0}^M$ and multi-view images $\{I_t^j\}_{j=0}^N$ at time $t$. The key idea of the solution is to utilize the performance predictions for image-branch pairs. The scheduler’s objective function is formulated as follows:
\begin{equation}
\begin{aligned}
c^* = \arg\max_c \textstyle \sum_{i=0}^M \sum_{j=0}^N f_A(B^i, I_t^j) \cdot d_{ij} \\
\text{s.t.} \quad \textstyle \sum_{i=0}^M \sum_{j=0}^N f_L(B^i) \cdot d_{ij} \leq T,
\end{aligned}
\label{equ:equation3}
\end{equation}
where $f_A$ and $f_L$ denote the branch accuracy and latency predictors, respectively. Here, $d_{ij}\in\{0,1\}$ indicates a binary decision, where $d_{ij}=1$ denotes the decision of $i^{\text{th}}$ branch to run for $j^{\text{th}}$ image. For each image, only one branch is allocated, i.e., $\textstyle \sum_{i=0}^M d_{ij} = 1, \forall j \in \{0,\ldots,N\}$. Otherwise, multiple images can be processed by a single branch, resulting $M^N$ possible configurations in total. In the following, we describe the method to find $c^*$.

\subsection{Performance Prediction}
\label{sec:performance_prediction}

To find the optimal selection of image-branch pairs, \system{} first predicts the expected accuracy and latency of each pair. As described in Section ~\ref{sec:preliminary}, the model’s detection capabilities are influenced by spatial characteristics, such as the number, location, and movement of objects. \system{} predicts changes in the spatial distribution considering these factors. The forecasted spatial distribution is then harnessed to predict the expected accuracy and processing time of possible branch-image selections.

\noindent\textbf{Prediction of spatial distribution.} To predict the expected spatial distribution at the incoming time $t$, Panopticus utilizes the future states of tracked objects estimated via a 3D Kalman, described in Section ~\ref{sec:multibranch_model_design}. Based on the objects' state predictions such as expected locations, each object is categorized considering all property levels described in Table ~\ref{tab:table2}. Consequently, objects are classified into one of 80 categories (5 distance levels $\times$ 4 velocity levels $\times$ 4 size levels). For instance, a pedestrian standing nearby has levels of D0, V0, and S0. The predicted spatial distribution vector $D_t^\prime$ is then calculated as follows:
\begin{equation}
D_t^\prime = [ p_{D0V0S0}, p_{D1V0S0}, \ldots, p_{D4V3S3} ],
\label{equ:equation4}
\end{equation}
where each element represents the ratio of the number of objects in each category to the total number of tracked objects at time $t$. In fact, \system{} calculates $D_t^{\prime j}$ for each camera view $j$. To do so, \system{} identifies the camera view that contains the predicted 3D center location of each object. Accordingly, the predicted distribution vectors allow for the relative comparison of expected scene complexities across camera views.

\noindent\textbf{Accuracy prediction.} The goal of the accuracy predictor $f_A$ is to predict the expected accuracy of each branch-image pair. We propose an approach to model the detection performance of the pairs by utilizing the predicted spatial distribution. We employ a regression model to realize the $f_A$. We use the XGBoost regressor for $f_A$, trained on a validation set from the nuScenes dataset. The purpose of $f_A$ is to predict the detection score (detailed in Section ~\ref{sec:eval_experiment_setup}) of each branch based on the spatial distribution vector. For each pair of detection branch and camera view $j$, $f_A$ takes the estimated spatial distribution $D_t^{\prime j}$ and a one-hot encoded branch type as inputs. As a result, the predicted scores for 16 detection branches are generated for each view $j$. To predict the detection score of the tracker's branch for each view $j$, $f_A$ utilizes $D_t^{\prime j}$ and the average confidence level of tracked objects. The rationale behind using confidence level is that the objects with higher certainty are more likely to appear in the near future.

\noindent\textbf{Latency prediction.} Latency prediction involves estimating the processing time of modules such as neural networks within detection branches. In general, these modules have consistent latency profiles at runtime. Accordingly, the expected latency of each detection branch can be determined simply by summing up its modules' latency profiles. Recall that forecasting the objects' future states can be performed instantaneously. On the other hand, the latency for updating the states of tracked objects by associating them with newly detected boxes depends on the number of objects in a given space. We modeled the processing time of state update with a simple linear regressor, trained by the same data used for $f_A$. The linear model is designed to predict the expected update latency as a function of the number of tracked objects.


\subsection{Execution Scheduling}
\label{sec:execution_scheduling}

Finding the optimal selection between multi-view images and inference branches can be solved using a combinatorial optimization solver. We use integer linear programming (ILP) because the image-branch selection is represented with a binary decision, and also our objective function and constraint are linear. Specifically, we adopt Simplex and branch-and-cut algorithms to efficiently find the optimal selection. The scheduling process including performance prediction takes up to 3ms on Orin Nano with limited computing power. Algorithm ~\ref{algorithm:algo1} shows the operational flow of our system which adapts to the surrounding space.

At a given time $t$, the system first estimates the spatial distributions for each camera view (line 2), based on the state predictions of tracked objects (line 11). Utilizing the spatial distributions, the system predicts the detection scores of all possible pairs of branches and camera views (line 3). To ensure a uniform comparison across camera views, each score is normalized against the score of the most powerful branch deployed on the target device. Next, the estimated latency for each branch is acquired using offline latency profiles (line 4). As our model includes modules that operate statically, such as the BEV head and tracker update, the effective latency limit is calculated by subtracting the estimated latencies of these modules from the latency target (lines 5-7). Then, the optimal selection of branch-image pairs is determined using the ILP solver (line 8). Taking the scheduler's decision and incoming images as inputs, the model generates 3D bounding boxes (lines 12-14). These outcomes are subsequently utilized to update the states of tracked objects (line 15). An outdated object without a matched detection box is penalized by halving the confidence level and removed if the confidence is lower than the threshold. Finally, a downstream application utilizes the information of detected objects to provide its functionality such as obstacle avoidance (line 16).

\begin{algorithm}[t]
\caption{Spatial-adaptive Execution Scheduling}
\fontsize{9pt}{9pt}\selectfont
\begin{algorithmic}[1]
\Statex \textbf{Initialize:} Object detector $O_D$ with branches $\{B^i\}_{i=0}^M$, Object tracker $O_T$, Accuracy predictor $f_A$, Latency predictor $f_L$, Latency target $T$, Integer linear programming solver ILP.
\Statex
\Function{Sched}{$X_t^\prime$}:
    \State $\{D_t^{\prime j}\}_{j=0}^N \gets$ calculate spatial distribution vectors from $X_t^\prime$
    \State $S^\prime \gets$ predict detection scores by $f_A(\{D_t^{\prime j}\}_{j=0}^N, \{B^i\}_{i=0}^M)$
    \State $L^\prime \gets$ get branch latencies by $f_L(\{B^i\}_{i=0}^M)$
    \State $L^\prime_{\text{update}} \gets$ predict tracker update time by $f_L(X_t^\prime)$
    \State $L_{\text{fixed}} \gets$ total profiled latency of fixed modules
    \State $T_{\text{MAX}} \gets T - L^\prime_{\text{update}} - L_{\text{fixed}}$
    \State $c^*_t \gets$ get optimal branch-image pairs by ILP($S^\prime$, $L^\prime$, $T_{\text{MAX}}$)
    \State \textbf{return} $c^*_t$
\EndFunction
\Statex
\Statex \# Initial frame is processed and objects are tracked by $O_T$
\While{system is running:}
    \State $X_t^\prime \gets$ get predicted states of tracked objects at time $t$
    \State $c^*_t \gets$ \Call{Sched}{$X_t^\prime$}
    \State $\{I_t^j\}_{j=0}^N \gets$ load multi-view images at time $t$
    \State $R_t \gets$ get detection results by $O_D(c_t^*, \{I_t^j\}_{j=0}^N)$
    \State Update tracked objects by associating them with $R_t$
    \State \Call{DownstreamApplication}{$R_t$}
\EndWhile
\end{algorithmic}
\label{algorithm:algo1}
\end{algorithm}

\section{Implementation}
\label{sec:implementation}

We implemented \system{} using Python and CUDA for GPU-based acceleration. All neural networks were developed using PyTorch~\cite{pytorch} and trained on the training set in the nuScenes dataset~\cite{nuscenes}. Note that \system{} is compatible with other 3D perception datasets such as Waymo~\cite{waymo_dataset}, having sensor configurations similar to nuScenes. The neural networks for 3D object detection are developed based on MMDetection3D~\cite{mmdet3d2020}. For the camera motion network, we customized and trained~\cite{visual-selective-vio} to produce consistent relative poses between two consecutive frames for any given camera view. For the object tracker, we modified SimpleTrack~\cite{simpletrack} to use velocity for 3D Kalman’s state transition model. We used the GPU-accelerated XGBoost library~\cite{xgboost} and linear model in scikit-learn~\cite{scikit-learn} for performance predictors, and PuLP~\cite{pulp} for the ILP solver. In the model adaptation stage, memory usage is profiled via tegrastats~\cite{tegrastats}.

We optimized the performance of our multi-branch model in various ways. We used TensorRT~\cite{tensorrt} to modularize the neural networks and to accelerate the inference. Networks converted to TensorRT are optimized using floating-point 16 (FP16) quantization and layer fusion, etc. Camera view images assigned to the same networks, such as backbone network or DepthNet, are batch-processed together. Additionally, we used CUDA Multi-Stream~\cite{cuda_streams_concurrency} to process these multiple networks in parallel on an edge GPU. Meanwhile, we noticed an accuracy loss due to the simplistic modularization of our multi-branch model. Specifically, DepthNets and BEV head trained with a specific backbone network are incompatible with others. One-size-fits-all DepthNet is not practical since each backbone generates a 2D feature map of different sizes. Thus, our model includes DepthNet variants tailored to each backbone, ensuring accurate depth estimation. In contrast, the BEV head takes inputs of a consistent shape, regardless of the preceding networks—backbones and DepthNets. This allows us to train a universal BEV head compatible with any combination of preceding networks. We trained the BEV head from scratch, while the pre-trained backbones and DepthNets were fine-tuned.

\section{Evaluation}
\label{sec:evaluation}

We evaluated the system’s accuracy and runtime efficiency through extensive experiments conducted on diverse edge devices and in real-world environments.

\subsection{Testbed and Dataset}
\label{sec:eval_testbed}

\begin{figure}[t]
    \centering
    \includegraphics[width=0.9\linewidth]{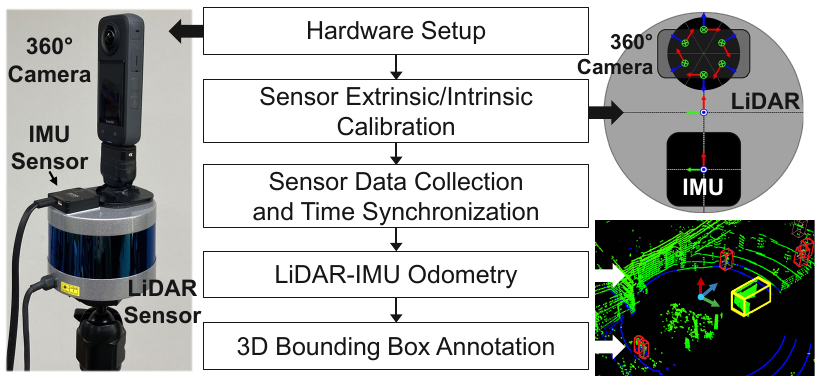}
    \vspace {-0.04in}
    \caption{Setup for mobile testbed and dataset.}
    \label{fig:figure9}
    \vspace {-0.2in}
\end{figure}

Omnidirectional 3D detection is crucial for various outdoor scenarios, such as robots navigating urban streets. However, only a public dataset for autonomous driving is available. To collect new datasets across diverse environments, such as campus and city square, we used our mobile 360° camera testbed. The datasets are summarized in Table ~\ref{tab:table3}, which we detail in the following.

\begin{table}[t]
\begin{minipage}{\columnwidth}
\centering
\caption{Datasets for the system evaluation.}
\vspace {-0.04in}
\label{tab:table3}
\begin{adjustbox}{width=\columnwidth,center}
\begin{tabular}{cccc}
\hline\hline
\multirow{2}{*}{\textbf{Place Category}} & \multirow{2}{*}{\textbf{\# of Scenes}} & \multirow{2}{*}{\shortstack{\textbf{\# of Frames} \\ \textbf{(\% Night)}}} & \multirow{2}{*}{\shortstack{\textbf{\# of Boxes} \\ \textbf{Per Frame}}} \\
& & & \\
\hline
Driving road~\cite{nuscenes} & 150 & 6,019 (11) & 0--119 \\
Campus & 8 & 1,027 (0) & 0--38 \\
Street & 6 & 1,006 (12) & 0--63 \\
Square/Crosswalk & 9 & 967 (28) & 0--54 \\
\hline\hline
\end{tabular}
\end{adjustbox}
\vspace{8pt} 
\caption{Edge devices used in the experiments.}
\vspace {-0.04in}
\label{tab:table4}
\begin{adjustbox}{width=\columnwidth,center}
\begin{tabular}{cccc}
\hline\hline
\textbf{Device} & \textbf{GPU} & \textbf{CPU} & \textbf{Memory} \\
\hline
\multirow{2}{*}{AGX Orin} & \multirow{2}{*}{\shortstack{2048-core Ampere \\ with 64 Tensor cores}} & \multirow{2}{*}{\shortstack{12-core Arm \\ A78AE v8.2}} & \multirow{2}{*}{31.3 GB} \\
& & & \\
\multirow{2}{*}{AGX Xavier} & \multirow{2}{*}{\shortstack{512-core Volta \\ with 64 Tensor cores}} & \multirow{2}{*}{\shortstack{8-core Carmel \\ Arm v8.2}} & \multirow{2}{*}{14.6 GB} \\
& & & \\
\multirow{2}{*}{Orin Nano} & \multirow{2}{*}{\shortstack{512-core Ampere \\ with 32 Tensor cores}} & \multirow{2}{*}{\shortstack{6-core Arm \\ A78AE v8.2}} & \multirow{2}{*}{6.3 GB} \\
& & & \\
\hline\hline
\end{tabular}
\end{adjustbox}
\end{minipage}
\vspace {-0.2in}
\end{table}

\begin{figure*}[t]
    \centering
    \begin{minipage}[t]{0.33\textwidth}
        \centering
        \includegraphics[width=\linewidth]{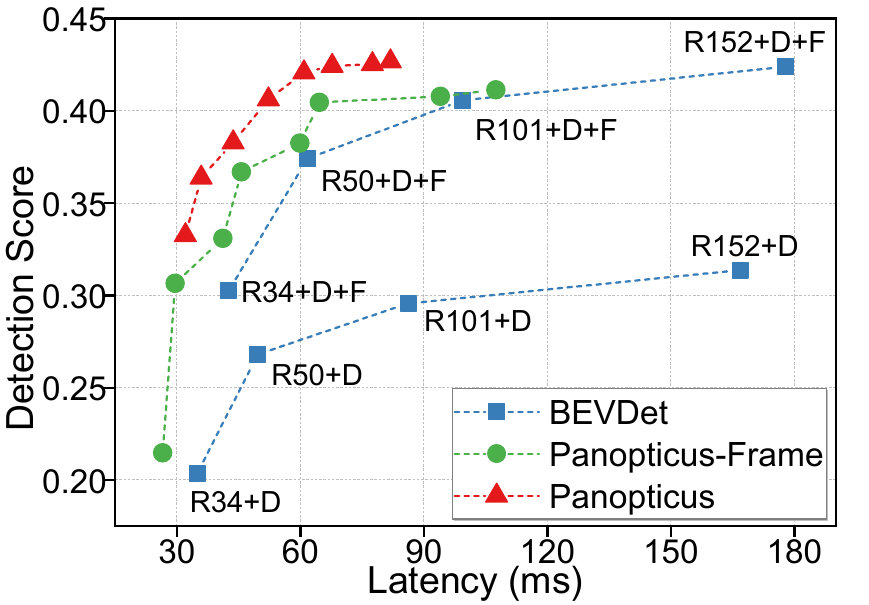}
        \subcaption{AGX Orin}
        \label{fig:figure10(a)}
    \end{minipage}
    \begin{minipage}[t]{0.33\textwidth}
        \centering
        \includegraphics[width=\linewidth]{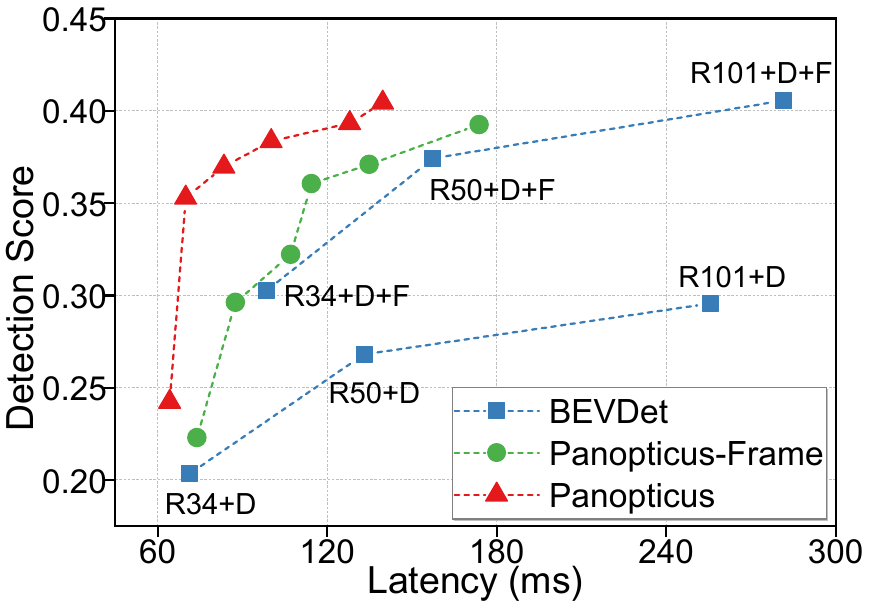}
        \subcaption{AGX Xavier}
        \label{fig:figure10(b)}
    \end{minipage}
    \begin{minipage}[t]{0.33\textwidth}
        \centering
        \includegraphics[width=\linewidth]{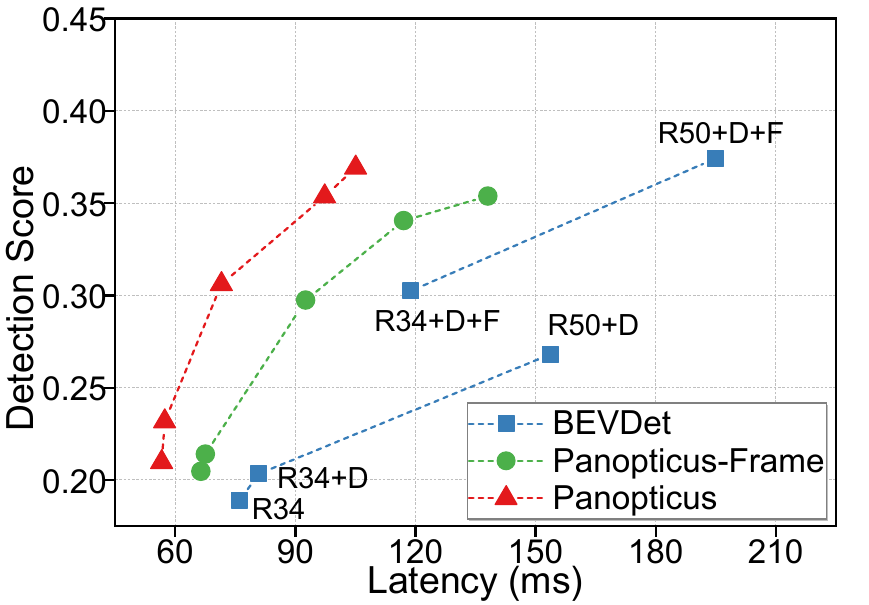}
        \subcaption{Orin Nano}
        \label{fig:figure10(c)}
    \end{minipage}
    \caption{Overall comparison of three edge devices using the nuScenes dataset. +D and +F denote the incorporation of dense DepthNet and temporal fusion, respectively.}
    \label{fig:figure10}
    \vspace{-0.2in}
\end{figure*}

For the public dataset, we used the nuScenes (described in Section ~\ref{sec:preliminary-setup}) with large-scale ground-truth annotations in urban driving scenes. We divided the total 850 scenes into sets of 600 for training, 100 for validation, and 150 for testing. Since the training and validation sets are used in our implementation, we conducted the experiments using a test set, which contains 6,019 labeled frames. We used six object types in the dataset—car, truck, bus, pedestrian, motorcycle, and bicycle—excluding less common types such as construction vehicle.

We built a mobile testbed based on a handheld 360° camera, as shown in Figure ~\ref{fig:figure9}. We configured hardware comprising a 360° camera (Insta360 X3~\cite{insta360x3}), LiDAR (Velodyne VLP-16~\cite{velodyne}), and IMU sensors. We calibrated the sensors' extrinsics to enable coordinate transformation among them. The 360° camera generates a 5.7K resolution rectangular image from two fisheye lenses. Due to severe distortion and rare object occurrence in the upper and lower areas, we excluded these areas. To mitigate the inherent distortion, we split the 360° image into six regions using perspective projection~\cite{tangent_image}. We calibrated each region’s intrinsics with the checkerboard method~\cite{zhang_method}, treating each as a separate virtual camera. From various outdoor spaces, such as campuses and squares, we collected sensor data and time-synchronized them. Accurate sensor trajectories were obtained using LiDAR-IMU odometry~\cite{pointlio}, which are then used to re-train the camera motion network. On each LiDAR point cloud, we use an AI-assisted tool~\cite{xtreme1} to manually label ground-truth 3D bounding boxes with the selected object types in nuScenes. Objects containing few LiDAR points, typically due to occlusion or distance, were excluded for annotation. Due to the labor-intensive nature of the labeling process, we selectively annotated 3,000 frames from 23 diverse and challenging scenes. Note that the neural networks for 3D detection are not re-trained with our 360° camera dataset, due to the lack of annotated data. However, our dataset contains the same object types in the nuScenes, ensuring compatibility.

\subsection{Experiment Setup}
\label{sec:eval_experiment_setup}

\textbf{Devices.} We used three Jetson edge devices with varying capabilities as summarized in Table ~\ref{tab:table4}. AGX Orin offers superior performance in terms of GPU/CPU compute power and memory capacity. Orin Nano has the least powerful GPU and memory but features a stronger CPU than AGX Xavier~\cite{jetson_xavier}, thanks to the hardware architecture differences.

\noindent\textbf{Baselines.} For the comparative analysis, we used various BEVDet variants described in our preliminary experiments as baselines. For a fair comparison, all baseline models were converted into FP16 TensorRT models like \system{} and used batched inference settings for multi-view images. We also used a modified version of \system{} as a baseline, named \systembaseline{}, allowing the inference branch to be switched \textit{per frame}, rather than \textit{per camera view} in each frame. Unlike \system{}, this baseline simply selects a branch of the highest predicted accuracy within the latency target, without an ILP solver. Note that even the concept of \systembaseline{} has not been proposed in prior works.

\noindent\textbf{Metrics.} As detailed in Section ~\ref{sec:preliminary-setup}, we used mAP and TP errors---mATE and mAVE---to assess the detection performance of our system and baselines. Prediction errors related to object size and orientation are not considered, as these errors remain consistently low across all evaluation targets. To evaluate our system and baselines using a single metric, we combine mAP and TP errors into a detection score (DS) similar to that in the nuScenes. It is calculated as a weighted sum of mAP, weighted by 6, and TP scores, each weighted by 2. Each TP score is acquired by $max(1-\text{TP error}, 0)$. The weighting scheme emphasizes the importance of detection accuracy while valuing the ability to predict objects’ location and speed.

\begin{figure}[t]
    \centering
    \includegraphics[width=\linewidth]{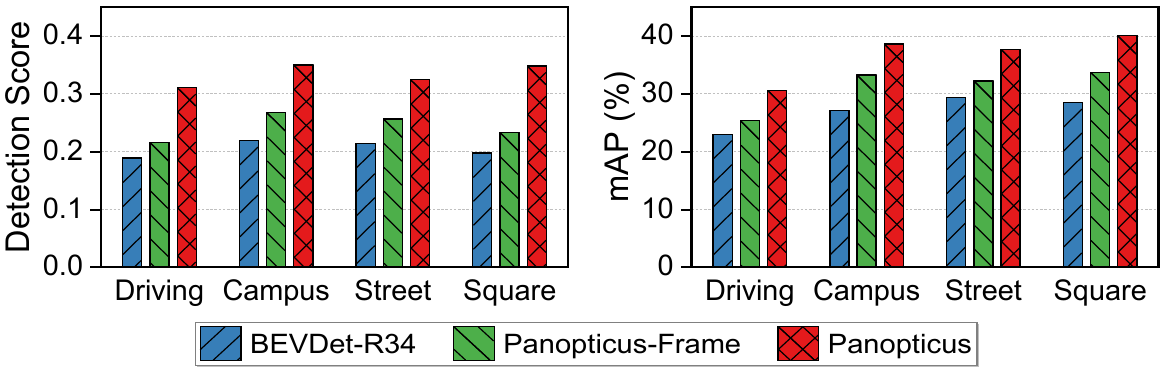}
    \caption{Performance comparison across scenes.}
    \label{fig:figure11}
    \vspace {-0.2in}
\end{figure}

\begin{figure*}[t]
    \centering
    \begin{minipage}[t]{0.49\textwidth}
        \centering
        \includegraphics[width=\textwidth]{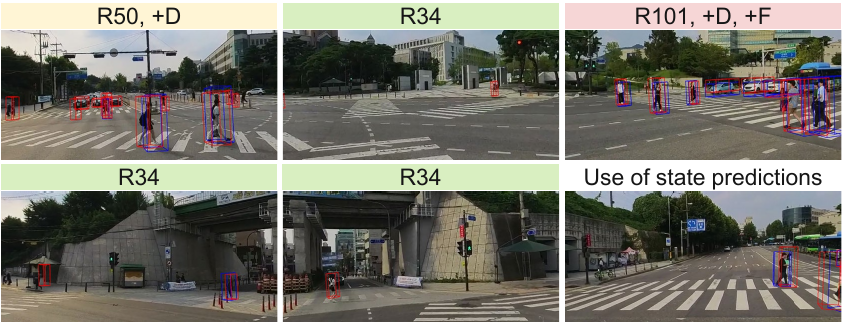}
        \vspace{-0.2in}
        \caption{A crosswalk scene in our dataset. Red and blue boxes indicate ground-truth and detected objects.}
        \label{fig:figure12}
    \end{minipage}
    \hfill
    \begin{minipage}[t]{0.49\textwidth}
        \centering
        \includegraphics[width=\textwidth]{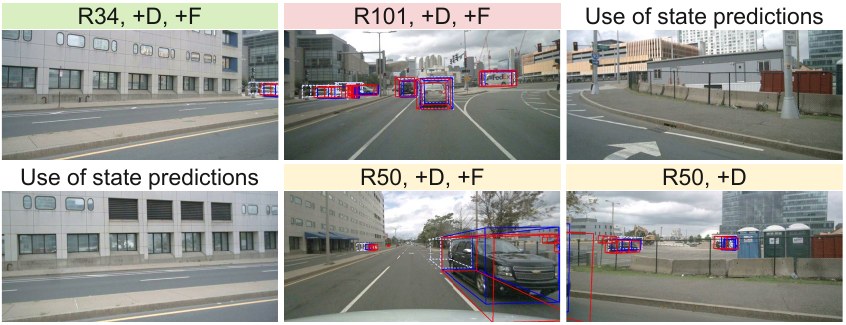}
        \vspace{-0.2in}
        \caption{A driving scene in nuScenes dataset. A box with dashed line indicates the previous state of an object.}
        \label{fig:figure13}
    \end{minipage}
    \vspace{-0.1in}
\end{figure*}

\begin{figure*}[t]
    \centering
    \begin{minipage}[t]{0.24\textwidth}
        \centering
        \includegraphics[width=\linewidth]{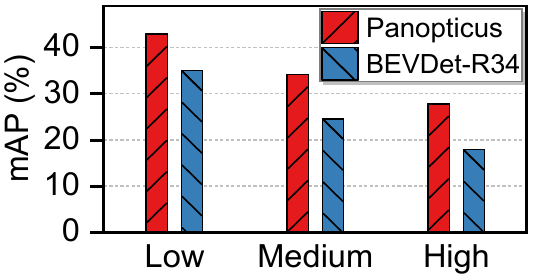}
        \subcaption{mAP over mean distance $/$ mean size levels}
        \label{fig:figure14(a)}
    \end{minipage}
    \hfill
    \begin{minipage}[t]{0.24\textwidth}
        \centering
        \includegraphics[width=\linewidth]{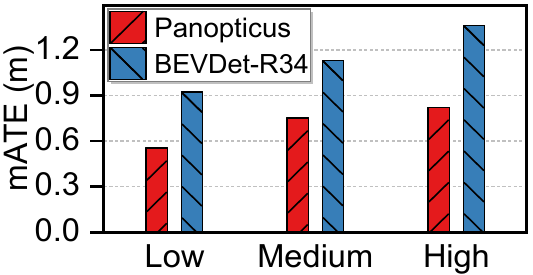}
        \subcaption{mATE over mean distance $/$ mean size levels}
        \label{fig:figure14(b)}
    \end{minipage}
    \hfill
    \begin{minipage}[t]{0.24\textwidth}
        \centering
        \includegraphics[width=\linewidth]{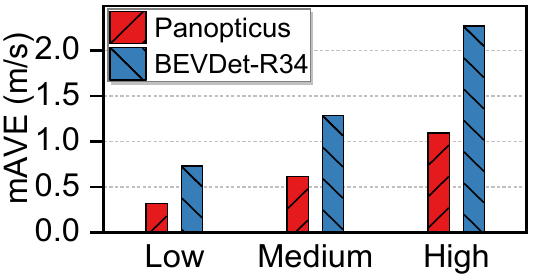}
        \subcaption{mAVE over mean velocity $/$ mean distance levels}
        \label{fig:figure14(c)}
    \end{minipage}
    \hfill
    \begin{minipage}[t]{0.2\textwidth}
        \centering
        \includegraphics[width=\linewidth]{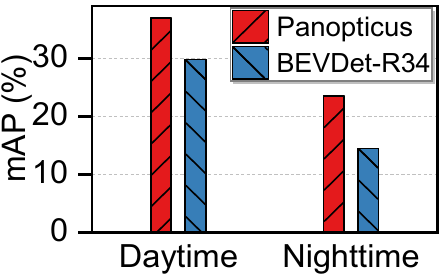}
        \subcaption{mAP over daytime vs. nighttime}
        \label{fig:figure14(d)}
    \end{minipage}
    \caption{System robustness across different scene complexities.}
    \label{fig:figure14}
    \vspace{-0.1in}
\end{figure*}

\subsection{Performance}
\label{sec:eval_performance}

\textbf{Overall comparison.} We compared the DS and latency of \system{} with its baselines across various devices and latency targets. For each experiment, we first launched the model adaptation process to accommodate the memory and latency constraints. For example, on Xavier, the R152 backbone and corresponding DepthNets, among other modules loaded on memory, are removed due to memory limits. For the fair comparison with baselines, we selected the inference times of baseline BEVDet models as the target latencies. We used the large-scale test set in the nuScenes for end-to-end performance comparison.

Figure ~\ref{fig:figure10} shows the overall DS and latency of our system and the baselines across all scenes in the test set. As shown in Figure ~\ref{fig:figure10(a)}, on Orin, \system{} achieved DS improvement of 41\% on average over baselines with similar latency. Notably, compared to the baseline model with the low inference time of 35ms, \system{} greatly improved DS by 79\%. Moreover, \system{} achieved latency reduction by 2.6$\times$ on average and up to 5.2$\times$, compared to the baseline models showing similar DS. The results demonstrate that \system{} efficiently utilizes edge computing resources by processing each camera view with a proper branch. The performance improvements are also observed in devices with much-limited computing power, as shown in Figure ~\ref{fig:figure10(b)} and ~\ref{fig:figure10(c)}. Compared to the baseline models with similar latency profiles, \system{} achieved average DS improvements of 51\% and 32\% on Xavier and Orin Nano, respectively. Also, on the same devices, the processing latencies are decreased by 2$\times$ and 1.7$\times$, respectively, on average compared to similar DS levels.

\systembaseline{} also achieved a performance gain by adaptively selecting a branch for each video frame. However, compared to \system{}, across all devices, \systembaseline{} has 7\% to 28\% lower DS, and the inference speed is 1.5$\times$ slower on average. The results showed that the coarse-grained level of branch selection leads to suboptimal performance. Processing all camera views using a computationally heavy branch may exceed the latency limits, thereby a lower-end branch is selected frequently.

\noindent\textbf{Comparison across different scenes.} We evaluated the performance of each scene category listed in Table ~\ref{tab:table3}. We used the AGX Orin for the experiments. We set a tight latency objective of 33ms, which corresponds to 30 frames per second (FPS), reflecting the safety-critical application scenarios such as obstacle avoidance. For the baseline models, we used BEVDet with an R34 backbone, which has a latency profile of 33ms on Orin. Figure ~\ref{fig:figure11} shows the comparison of DS and mAP on each scene type. In the case of average DS across all scenes, \system{} outperformed the baseline model and \systembaseline{} by 62\% and 38\%, respectively, under 30 FPS constraints. The average mAP improvement was 36\% over the baseline model and 18\% over the \systembaseline{}. Due to the complex and dynamic nature of the road environments, mAP is relatively decreased for the driving scenes.


\subsection{Robustness}
\label{sec:eval_robustness}

\system{} adjusts its operation based on the prediction of spatial distribution, considering the diverse properties of surrounding objects. We analyzed the robustness of \system{} under various circumstances. The experiments were conducted using Orin under real-time condition of 30 FPS. As a baseline meeting the condition, we used BEVDet with the R34 backbone.

\noindent\textbf{Impact of object distance and size.} We analyzed how the detection capability is influenced by the distance (from the camera) and size of the object. Experiments were conducted on our mobile 360° camera dataset, which has diversity in object distance and size. For each video frame, we first obtained the distance ($m$) and size ($m^3$) of all objects and calculated the mean of distance and size. Then, we defined the scene complexity of the frame as the value of mean distance divided by the mean size, classified into one of three complexity levels. Figure ~\ref{fig:figure14(a)} shows the changes in mAP as the complexity level increases. The results showed that mAP decreases as the level increases---the negative impact is significant when a large portion of objects are distant and small, which appear less discernible in images. Compared to the baseline model without considering such spatial complexity, \system{} successfully enhanced the accuracy at a \textit{High} level by 56\%. This effect is also observed in the prediction error of the object’s 3D location, as shown in Figure ~\ref{fig:figure14(b)}. \system{} reduced mATE on \textit{High} complexity level of frames by 40\% compared to the baseline model. Figure ~\ref{fig:figure12} displays an example frame from our mobile camera dataset, showing the branch selection of our scheduler and resulting detection boxes. \system{} allocated more resources to views with many distant or small objects, processing them using enhanced image feature extraction and depth estimation.

\noindent\textbf{Impact of object distance and velocity.} We analyzed the impact of object distance and velocity, particularly in dynamic driving scenes where objects often move quickly. We set the scene complexity level of each frame based on the ratio of the objects’ average velocity ($m/s$) to the average distance. We found that the object speed barely impacts mAP. However, as shown in Figure ~\ref{fig:figure14(c)}, the velocity prediction error, i.e., mAVE, greatly increases followed by an increasing number of objects moving fast at a close distance. This is because fast and proximate objects present large positional shifts between two video frames, as seen in Figure ~\ref{fig:figure13}, making speed prediction challenging. \system{} processes camera views containing such objects by fusing two consecutive BEV feature maps. As a result, \system{} reduced mAVE by 2.2$\times$ on average compared to the baseline.

\noindent\textbf{Impact of time of day.} To compare the detection performance across different times of the day, we categorized all scenes from the datasets into two groups: daytime and nighttime. As shown in Figure ~\ref{fig:figure14(d)}, mAP decreases during nighttime, which is attributed to motion blur in images captured under low lighting conditions at night. Notably, in nighttime scenes, \system{} achieved a 63\% improvement in mAP compared to the baseline, showing its robustness to varying lighting conditions.

\subsection{Component Analysis}
\label{sec:eval_component_analysis}


\begin{figure}[t]
    \centering
    \begin{minipage}[t]{0.46\linewidth}
        \centering
        \includegraphics[width=0.85\linewidth]{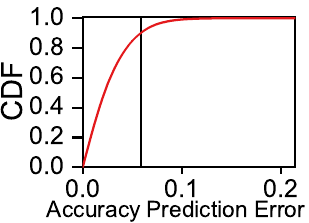}
        \subcaption{CDF of DS pred. error}
        \label{fig:figure15(a)}
    \end{minipage}
    \hspace{0.01\linewidth} 
    \begin{minipage}[t]{0.5\linewidth}
        \centering
        \includegraphics[width=0.85\linewidth]{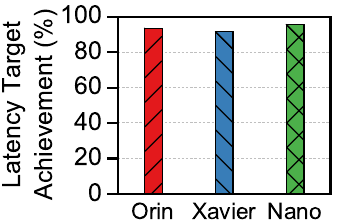}
        \subcaption{Latency target compliance}
        \label{fig:figure15(b)}
    \end{minipage}
    \begin{minipage}[t]{0.46\linewidth}
        \centering
        \includegraphics[width=0.85\linewidth]{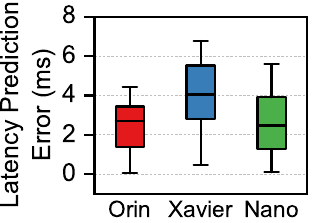}
        \subcaption{Latency prediction error}
        \label{fig:figure15(c)}
    \end{minipage}
    \hspace{0.01\linewidth} 
    \begin{minipage}[t]{0.5\linewidth}
        \centering
        \includegraphics[width=0.85\linewidth]{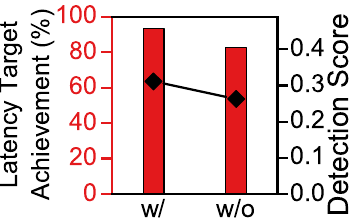}
        \subcaption{Effect of tracker’s branch}
        \label{fig:figure15(d)}
    \end{minipage}
    \caption{Evaluation of various system components.}
    \label{fig:figure15}
    \vspace{-0.2in}
\end{figure}

We evaluated the performance of the system component of \system{}. Experiments were conducted using the test set in the nuScenes.

\noindent\textbf{Accuracy predictor.} \system{} predicts the accuracy (i.e., DS) based on the expected spatial distribution to assign appropriate branches to different camera views. Figure ~\ref{fig:figure15(a)}, showing the CDF of DS prediction errors, indicates that the 90th percentile error is less than 0.06. Given that \system{} selects optimal branches based on the relative difference in accuracy predictions across camera views, the negative impact on performance from these errors is negligible.

\noindent\textbf{Latency predictor.} Our scheduler selects optimal inference branches based on the predicted latencies of the system’s modules. Accurate latency prediction is crucial to meet the latency constraints. Figure ~\ref{fig:figure15(b)} shows how many frames were processed within the tight latency targets of 35ms, 70ms, and 80ms on Orin, Xavier, and Orin Nano, respectively. These targets correspond to the inference times of the lowest-performing baseline models shown in Figure ~\ref{fig:figure10}. On average, ~\system{} achieved 94\% latency satisfaction across all devices, showing the robustness of the latency predictor. We observed, as shown in Figure ~\ref{fig:figure15(c)}, that the prediction error between the expected latency and the actual latency varied across devices. We identified that the increase in errors on Xavier is due to the relatively large variation in the processing time of the tracker’s state update, which is caused by a low-performance CPU. This characteristic led to an increase in prediction errors of state update latencies. Across all devices, however, we observed that the actual latencies of the system mostly fall below the target latencies. Therefore, the impact of latency prediction errors is minimized to meet the constraints.

\noindent\textbf{Gain with tracker’s branch.} \system{} includes a lightweight branch that outputs the predicted states of tracked objects for the target camera view, thereby skipping the detection. We conducted an ablation study to verify the gain of using the tracker’s branch. As shown in Figure ~\ref{fig:figure15(d)}, without (w/o) the tracker’s branch, the DS and latency target satisfaction are reduced by 13\% and 18\%, respectively, compared to with (w/) the branch. This highlights the importance of using the lightweight branch to effectively balance the detection accuracy and efficiency.

\begin{table}[t]
\caption{System overhead.}
\label{tab:table5}
\centering
\begin{adjustbox}{width=0.9\columnwidth,center}
\begin{tabular}{cccc}
\hline\hline
Latency Target (ms) & Memory (GB) & CPU (W) & GPU (W) \\
\hline
33 & 15.5 & 3.3 & 6.3 \\
150 & 16.7 & 3.6 & 8.1 \\
\hline\hline
\end{tabular}
\end{adjustbox}
\vspace {-0.2in}
\end{table}

\subsection{Overhead}
\label{sec:eval_overhead}

We analyzed the memory and power consumption of \system{}. For the experiment, we obtained the runtime traces of memory and power consumption using tegrastats~\cite{tegrastats}. The traces were collected every 100ms while executing the system on Orin. We compared the overheads under different latency constraints, 33ms and 150ms. As shown in Table ~\ref{tab:table5}, \system{} consumed 15.5 to 16.7 GB of memory, which is around half of Orin’s memory limit. A large portion of memory overheads is due to the intermediate buffers used for inference acceleration with TensorRT~\cite{ibuilderconfig}. Under the latency target of 150ms, average and peak power consumption of GPU is 8.1W and 16.7W, respectively. An increased memory and power consumption under the target of 150ms is due to the frequent utilization of powerful branches with large-scale DNN modules.

\section{Related Work}
\label{sec:relwork}

\textbf{Adaptive object detection systems.} Many computer vision systems designed for resource-constrained edge devices have achieved efficient resource utilization by adapting to the video content or computing budgets~\cite{mobicom2018nestdnn, dac2019reform, sec2020flexdnn, mobicom2021legodnn, mobicom2022neulens}. In parallel with these advancements, several works have specifically focused on adaptive object detection methods~\cite{chin2019adascale, remix, approxdet}, facilitating valuable applications such as edge video analytics or mobile augmented reality. For instance, ApproxDet~\cite{approxdet} presented a multi-branch framework that switches between a detector and a tracker based on awareness of video content characteristics and resource contention. Remix~\cite{remix} brings video content adaptiveness by partitioning images and selectively applying neural networks with different scales. However, existing adaptive methods fall short in supporting resource-efficient omnidirectional 3D detection. This type of detection requires processing each camera view using different 3D perception capabilities, such as enhanced estimation of objects' 3D location and velocity. To process each view optimally, \system{} predicts its expected performance based on short-term future dynamics in the spatial distribution, which is crucial for mobile scenarios. Although Remix~\cite{remix} utilizes performance estimation based on object distribution, it relies on long-term historical distribution and does not account for various object characteristics in 3D, which is unsuitable for our system's setting.

\noindent\textbf{3D object detection on edge devices.} With the rapid advancement of edge computing, there is an increasing demand for employing 3D object detection on edge devices. DeepMix~\cite{DeepMix} addressed the limitations of edge resources by delegating the compute-intensive 2D detection tasks to a server equipped with high-performance GPUs. The other lightweight tasks, such as estimating the 3D location of detected objects using a depth sensor, are efficiently handled by mobile devices. Another solution, PointSplit~\cite{PointSplit}, proposed a parallel processing technique that utilizes edge NPU and GPU to facilitate on-device execution of RGB-D camera-based 3D detection. This approach exemplifies the trend of harnessing the power of specialized AI accelerators to meet the demands of edge computing~\cite{band_mobisys, zhao2023miriam, blastnet_sensys_22, codl_mobisys22}. VIPS~\cite{vips_mobicom}, designed for self-driving vehicles, introduced an edge-based system that collaborates with outdoor infrastructures equipped with computing units and LiDAR sensors. This strategy effectively extended the vehicles' perception ranges by fusing data from the onboard system and the infrastructure. Apart from these efforts, \system{} effectively initiates a self-contained, comprehensive perception system for resource-constrained edge devices. The system introduced a novel concept of camera-based omnidirectional 3D perception optimized for edge computing abilities, eliminating the need for depth sensors or computation offloading.


\section{Discussion and Future Work}
\label{sec:discussion}

\noindent\textbf{Supporting multiple tasks.} As \system{} is the initial endeavor toward adaptive omnidirectional 3D detection, our current implementation operates under the assumption of a single-task execution environment. Within this environment, the system's performance characteristics, such as offline latency profiles of the multi-branch model, are consistent during runtime. However, in real-world applications like mobile robot navigation, it is common to run 3D object detection alongside other critical tasks, such as odometry or path planning. The future improvements of \system{} considering the multi-task execution environment could be achieved in several ways. First, monitoring the runtime dynamics caused by resource contention among concurrent tasks is crucial for optimized resource utilization. Second, there is a need to explore how to co-design multi-branch models for different 3D tasks and optimize those models given the constraints of application and device. Lastly, the better utilization of available heterogeneous processors, such as CPUs and NPUs, could enhance the efficiency of the multi-task workload.

\noindent\textbf{Selection criteria for performance metrics.} For the design and evaluation of \system{}, we used the most popular 3D detection performance metrics---detection score~\cite{nuscenes} and mAP. These metrics are effective in assessing the overall performance of a 3D detection system. However, the system's effectiveness could be further improved through metric design that considers application-specific requirements. For instance, detecting fast and close objects is crucial for the safety of robot navigation systems, as opposed to detecting slowly moving or distant objects. Also, the importance of different object types varies depending on application scenarios. This application-centric metric design is worth exploring for future improvements.


\section{Conclusion}
\label{sec:conclusion}

This paper proposed \system{}, an omnidirectional and camera-based 3D object detection system designed for resource-constrained edge devices. \system{} effectively balances detection accuracy and latency by employing a multi-branch model that selects the optimal inference configuration for each camera view based on predicted spatial characteristics. Extensive experiments have shown that \system{} outperforms its baselines across various environments and edge devices, highlighting its potential to enhance applications requiring real-time perception of surrounding 3D objects.


\begin{acks}
This work was supported by the National Research Foundation of Korea (NRF) grant
funded by the Korea government (MSIT) (No. RS-2024-00344323).
\end{acks}

\balance
\bibliographystyle{acmart/ACM-Reference-Format}
\bibliography{references}

\end{document}